\documentclass{article}
\PassOptionsToPackage{numbers,sort,compress}{natbib}

\usepackage[preprint]{neurips_2019}

\usepackage{natbib}
\usepackage[utf8]{inputenc} 
\usepackage[T1]{fontenc}    
\usepackage{hyperref}       
\hypersetup{colorlinks,linkcolor={blue},citecolor={blue},urlcolor={red}}  
\usepackage{url}            
\usepackage{booktabs}       
\usepackage{amsfonts}       
\usepackage{times}
\usepackage{epsfig}
\usepackage{graphicx}
\usepackage{amsmath}
\usepackage{amssymb}
\usepackage{mathrsfs}  
\usepackage{multirow}
\usepackage[linesnumbered, ruled, vlined]{algorithm2e}
\usepackage{nicefrac}       
\usepackage{microtype}      

\newcommand{\cmtt}[1]{{\fontfamily{cmtt}\selectfont #1}}

\begin{document}

\title{Fast AutoAugment}

\author{
    Sungbin Lim\thanks{Equal Contribution} \\
    Kakao Brain \\
    \cmtt{sungbin.lim@kakaobrain.com} \\
    \And
    Ildoo Kim\footnotemark[1] \\
    Kakao Brain \\
    \cmtt{ildoo.kim@kakaobrain.com} \\
    \And
    Taesup Kim \\
    MILA, Universit\'e de Montr\'eal, Canada \\
    \cmtt{taesup.kim@umontreal.ca}    
    \And
    Chiheon Kim \\
    Kakao Brain \\
    \cmtt{chiheon.kim@kakaobrain.com} \\
    \And
    Sungwoong Kim \\
    Kakao Brain \\
    \cmtt{swkim@kakaobrain.com} \\
}

\maketitle


\begin{abstract}
Data augmentation is an essential technique for improving generalization ability of deep learning models. Recently, AutoAugment \cite{cubuk2018autoaugment} has been proposed as an algorithm to automatically search for augmentation policies from a dataset and has significantly enhanced performances on many image recognition tasks. However, its search method requires thousands of GPU hours even for a relatively small dataset. In this paper, we propose an algorithm called Fast AutoAugment that finds effective augmentation policies via a more efficient search strategy based on density matching. In comparison to AutoAugment, the proposed algorithm speeds up the search time by orders of magnitude while achieves comparable performances on image recognition tasks with various models and datasets including CIFAR-10, CIFAR-100, SVHN, and ImageNet.
\end{abstract}

\section{Introduction}

Deep learning has become a state-of-the-art technique for computer vision tasks, including object recognition \cite{yamada2018shakedrop, real2018regularized, hu2018squeeze}, detection \cite{ren2015faster, liu2016ssd}, and segmentation \cite{chen2018deeplab, he2017mask}. However, deep learning models with large capacity often suffer from overfitting unless significantly large amounts of labeled data are supported. Data augmentation (DA) has been shown as a useful regularization technique to increase both the quantity and the diversity of training data. Notably, applying a carefully designed set of augmentations rather than naive random transformations in training improves the generalization ability of a network significantly \cite{krizhevsky2012imagenet, paschali2019data}. However, in most cases, designing such augmentations has relied on human experts with prior knowledge on the dataset.
  
With the recent advancement of automated machine learning (AutoML), there exist some efforts for designing an automated process of searching for augmentation strategies directly from a dataset. AutoAugment \cite{cubuk2018autoaugment} uses reinforcement learning (RL) to automatically find data augmentation policy when a target dataset and a model are given. It samples an augmentation policy at a time using a controller RNN, trains the model using the policy, and gets the validation accuracy as a reward to update the controller. AutoAugment especially achieves a dramatic improvement in performances on several image recognition benchmarks. However, AutoAugment requires thousands of GPU hours even in a reduced setting, in which the size of the target dataset and the network is small. Recently proposed Population Based Augmentation (PBA) \cite{ho2019pba} is a method to deal with this problem, which is based on population-based training method of hyperparameter optimization. In contrast to previous methods, we propose a new search strategy that does not require any repeated training of child models. Instead, the proposed algorithm directly searches for augmentation policies that maximize the match between the distribution of augmented split and the distribution of another, unaugmented split via a single model.

\begin{table}[t!] \center
\label{table:GPU hours}
\begin{tabular}{c c c c }
\toprule
Dataset & AutoAugment \cite{cubuk2018autoaugment} & Fast AutoAugment
\tabularnewline
\midrule
CIFAR-10 & 5000 & 3.5 \\
SVHN & 1000 & 1.5 \\
ImageNet & 15000 & 450
\tabularnewline
\bottomrule
\end{tabular}
\caption{GPU hours comparison of Fast AutoAugment with AutoAugment. We estimate computation cost with an NVIDIA Tesla V100 while AutoAugment measured computation cost in Tesla P100.}
\end{table}

In this paper, we propose an efficient search method of augmentation policies, called Fast AutoAugment, motivated by Bayesian DA \cite{tran2017bayesian}. Our strategy is to improve the generalization performance of a given network by learning the augmentation policies which treat augmented data as missing data points of training data. However, different from Bayesian DA, the proposed method recovers those missing data points by the exploitation-and-exploration of a family of inference-time augmentations \cite{simonyan2014very, szegedy2016rethinking} via Bayesian optimization in the policy search phase. We realize this by using an efficient density matching algorithm that does not require any back-propagation for network training for each policy evaluation. The proposed algorithm can be easily implemented by making good use of distributed learning frameworks such as \cmtt{Ray} \cite{ray}.

Our experiments show that the proposed method can search augmentation policies significantly faster than AutoAugment (see Table \ref{table:GPU hours}), while retaining comparable performances to AutoAugment on diverse image datasets and networks, especially in two use cases: (a) direct augmentation search on the dataset of interest, (b) transferring learned augmentation policies to new datasets. On ImageNet, we achieve an error rate of 19.4\% for ResNet-200 trained with our searched policy, which is 0.6\% better than 20.0\% with AutoAugment.

This paper is organized as follows. First, we introduce related works on automatic data augmentation in Section \ref{sec:related-work}. Then, we present our problem setting to achieve the desired goal and suggest Fast AutoAugment algorithm to solve the objective efficiently in Section \ref{sec:faa}. Finally, we demonstrate the efficiency of our method through comparison with baseline augmentation methods and AutoAugment in Section \ref{sec:experiment}.

\section{Related Work}
\label{sec:related-work}
There are many studies on data augmentation, especially for image
recognition. On the benchmark image dataset, such as CIFAR and ImageNet, random crop, flip, rotation, scaling, and color transformation, have been performed as baseline augmentation methods \cite{krizhevsky2012imagenet,Han_2017_CVPR,sato2015apac}. Mixup \cite{zhang2017mixup}, Cutout \cite{devries2017cutout}, and CutMix \cite{yun2019cutmix} have been recently proposed to either replace or mask out the image patches randomly and obtained more improved performances on image recognition tasks. However, these methods are designed manually based on domain knowledge.

Naturally, automatically finding data augmentation methods from data in principle has emerged to overcome the performance limitation that originated from a cumbersome exploration of methods by a human. Smart Augmentation \cite{lemley2017smart} introduced a network that learns to generate augmented data by merging two or more samples in the same class. \cite{shrivastava2017learning} employed a generative adversarial network (GAN) \cite{goodfellow2014generative} to generate images that augment datasets. Bayesian DA \cite{tran2017bayesian} combined Monte Carlo expectation maximization algorithm with GAN to generate data by treating augmented data as missing data points on the distribution of the training set. 

Due to the remarkable successes of NAS algorithms on various computer vision tasks \cite{real2018regularized, zoph2018learning, kim2018scalable}, several current studies also deal with automated search algorithms to obtain augmentation policies for given datasets and models. The main difference between the previously learned methods and these automated augmentation search methods is that the former methods exploit generative models to create augmented data directly, whereas the latter methods find optimal combinations of predefined transformation functions. AutoAugment \cite{cubuk2018autoaugment} introduced an RL based search strategy that alternately trained a child model and RNN controller and showed the state-of-the-art performances on various datasets with different models. Recently, PBA \cite{ho2019pba} proposed a new algorithm which generates augmentation policy schedules based on population based training \cite{jaderberg2017population}. Similar to PBA, our method also employs hyperparameter optimization to search for optimal policies but uses Tree-structured Parzen Estimator (TPE) algorithm \cite{bergstra2011algorithms} for practical implementation.

\section{Fast AutoAugment}
\label{sec:faa}

In this section, we first introduce the search space of the symbolic augmentation operations and formulate a new search strategy, efficient density matching, to find the optimal augmentation policies efficiently. We then describe our implementation based on Bayesian hyperparameter optimization incorporated into a distributed learning framework.

\subsection{Search Space}
\label{subsec:search-space}

 \begin{figure}
     \centering
     \includegraphics[scale=0.38]{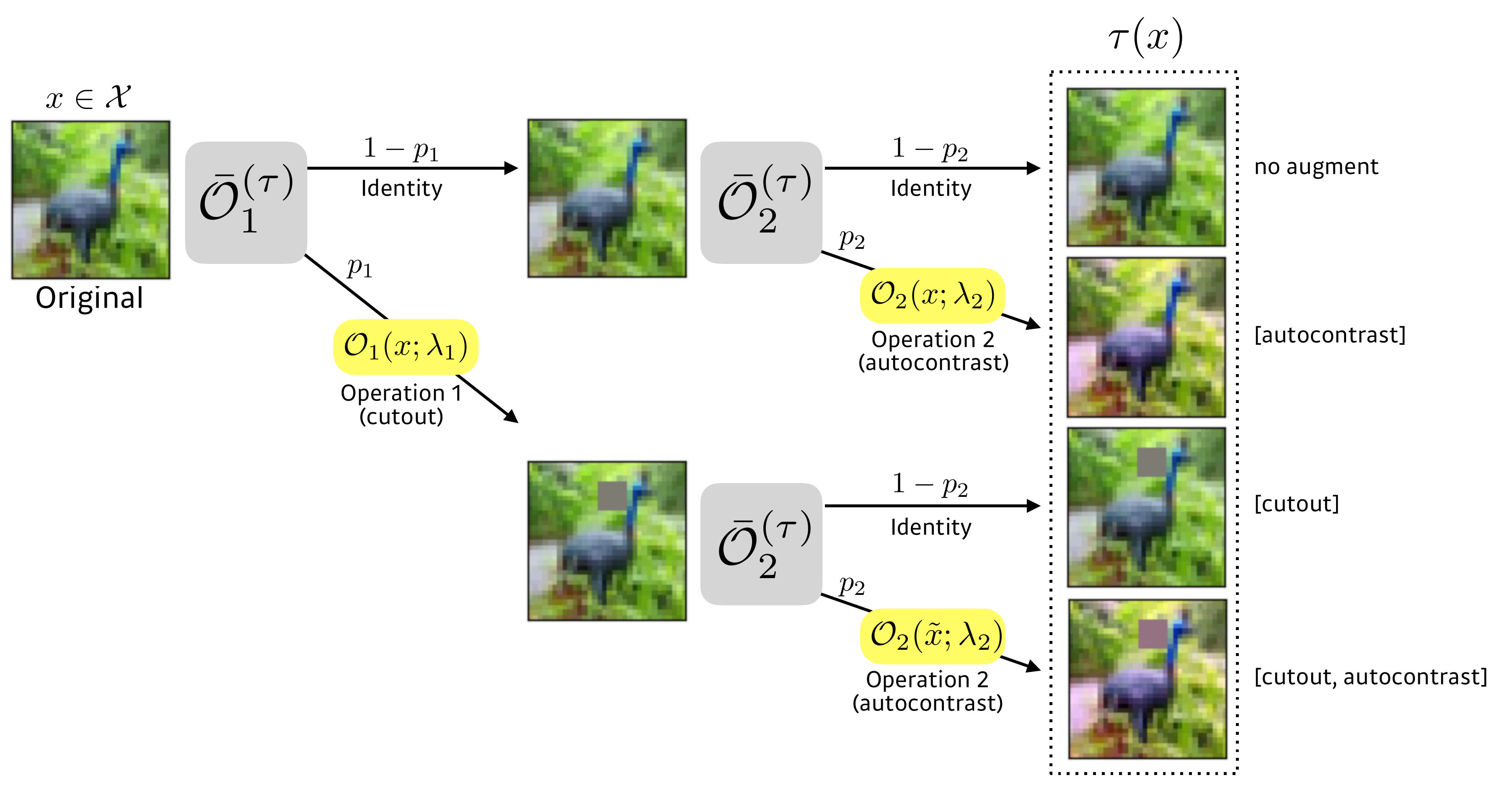}

     \caption{An example of augmented images via a sub-policy in the search space $\mathcal{S}$. Each sub-policy $\tau$ consists of 2 operations; for instance, $\tau=$[\cmtt{cutout}, \cmtt{autocontrast}] is used in this figure. Each operation $\bar{\mathcal{O}}_{i}^{(\tau)}$ has two parameters: the probability $p_{i}$ of calling the operation and the magnitude $\lambda_{i}$ of the operation. These operations are applied with the corresponding probabilities. As a result, a sub-policy randomly maps an input data to the one of 4 images. Note that the identity map (no augmentation) is also possible with probability $(1-p_{1})(1-p_{2})$.}
     \label{fig:sub-policy}
 \end{figure}

Let $\mathbb{O}$ be a set of augmentation (image transformation) operations $\mathcal{O}:\mathcal{X}\to\mathcal{X}$ defined on the input image space $\mathcal{X}$. Each operation $\mathcal{O}$ has two parameters: the calling probability $p$ and the magnitude $\lambda$ which determines the variability of operation. Some operations (e.g. \cmtt{invert}, \cmtt{flip}) do not use the magnitude. Let $\mathcal{S}$ be the set of sub-policies where a sub-policy $\tau\in\mathcal{S}$ consists of $N_{\tau}$ consecutive operations $\{\bar{\mathcal{O}}_{n}^{(\tau)}(x; p_{n}^{(\tau)},\lambda_{n}^{(\tau)}):n=1,\ldots,N_{\tau}\}$ where each operation is applied to an input image sequentially with the probability $p$ as follows:
\begin{align}
    \label{eq:operation}
\bar{\mathcal{O}}(x; p,\lambda):=\begin{cases}
\mathcal{O}(x; \lambda) & :\text{with probability }p\\
x & :\text{with probability }1-p.
\end{cases}    
\end{align}
Hence, the output of sub-policy $\tau(x)$ can be described by a composition of operations as 
$$
\tilde{x}_{(n)}=\bar{\mathcal{O}}_{n}^{(\tau)}(\tilde{x}_{(n-1)}),\quad n=1,\ldots,N_{\tau}
$$
where $\tilde{x}_{(0)}=x$ and $\tilde{x}_{(N_{\tau})}=\tau(x)$. Figure \ref{fig:sub-policy} shows a specific example of augmented images by $\tau$. Note that each sub-policy $\tau$ is a random sequence of image transformations which depend on $p$ and $\lambda$, and this enables to cover a wide range of data augmentations. Our final policy 
$\mathcal{T}$ is a collection of $N_{\mathcal{T}}$ sub-policies and $\mathcal{T}(D)$
indicates a set of augmented images of dataset $D$ transformed by every sub-policies $\tau\in\mathcal{T}$:
$$
\mathcal{T}(D) = \bigcup_{\tau\in\mathcal{T}}\{(\tau(x),y):(x,y)\in D\}
$$
Our search space is similar to previous methods except that we use both continuous values of probability $p$ and magnitude $\lambda$ at $[0,1]$ which has more possibilities than discretized search space. 


\subsection{Search Strategy}
\label{subsec:search-strategy}

In Fast AutoAugment, we consider searching the augmentation policy as a density matching between a pair of train datasets. Let $\mathcal{D}$ be a probability distribution on $\mathcal{X}\times\mathcal{Y}$
and assume dataset $D$ is sampled from this distribution. For a given classification model $\mathcal{M}(\cdot|\theta):\mathcal{X}\to\mathcal{Y}$ that is parameterized by $\theta$, the expected accuracy and the expected loss of $\mathcal{M}(\cdot|\theta)$  
on dataset $D$ are denoted by $\mathcal{R}(\theta|D)$ and $\mathcal{L}(\theta|D)$, respectively.

\subsubsection{Efficient Density Matching for Augmentation Policy  Search}
\label{subsubsec:FAA}

 \begin{figure}
     \centering
     \includegraphics[width=\textwidth]{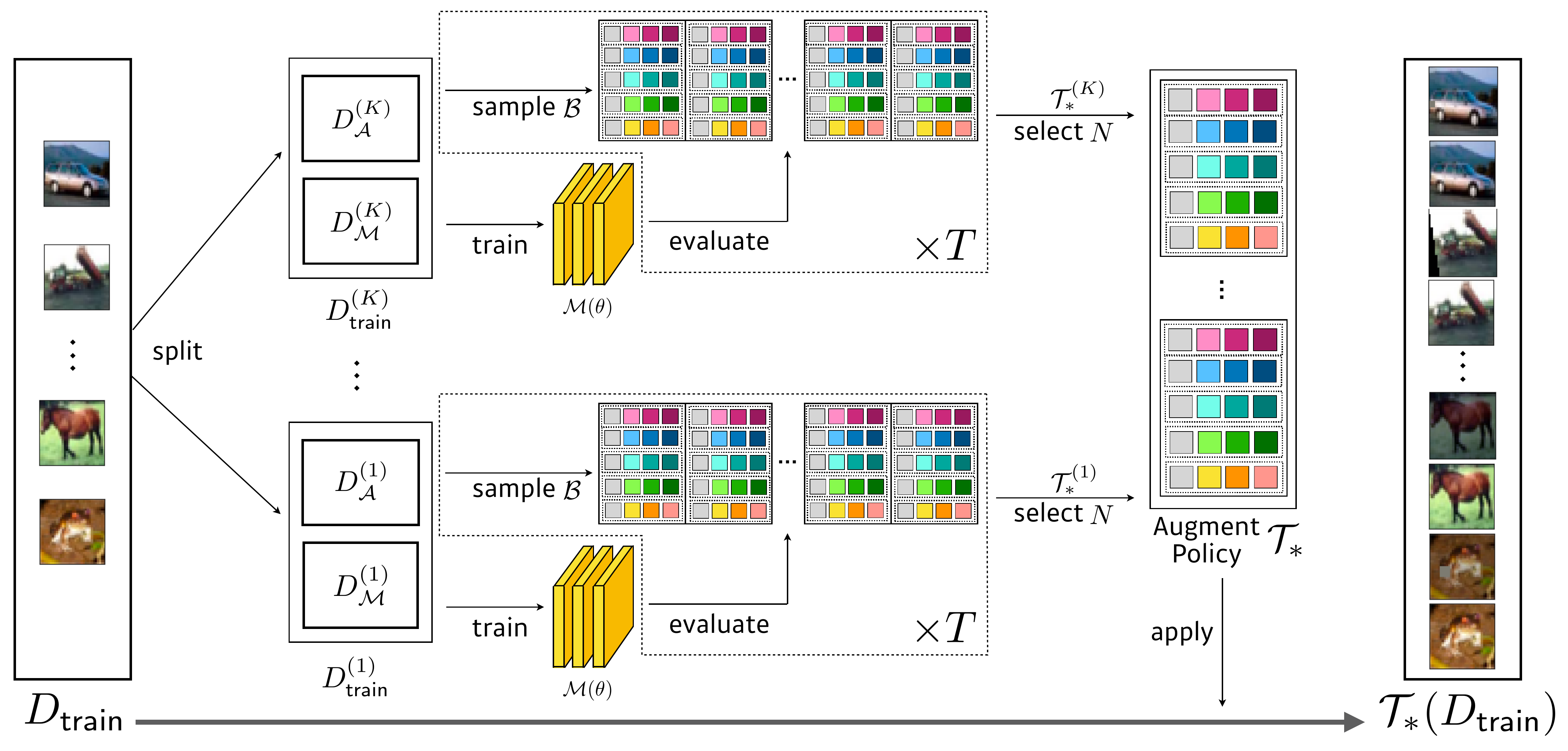}

     \caption{An overall procedure of augmentation search by Fast AutoAugment algorithm. For exploration, the proposed method splits the train dataset $D_{\text{train}}$ into $K$-folds, which consists of two datasets $D_{\mathcal{M}}^{(k)}$ and $D_{\mathcal{A}}^{(k)}$. Then model parameter $\theta$ is trained in parallel on each $D_{\mathcal{M}}^{(k)}$. After training $\theta$, the algorithm evaluates $B$ bundles of augmentation policies on $D_{\mathcal{A}}$ without training $\theta$. The top-$N$ policies obtained from each $K$-fold are appended to an augmentation list $\mathcal{T}_{*}$.}
     \label{fig:FAA}
 \end{figure}

For any given pair of $D_{\text{train}}$ and $D_{\text{valid}}$, our goal is to improve the generalization ability by searching the augmentation policies that match the density of $D_{\text{train}}$ with density of augmented $D_{\text{valid}}$. However, it is impractical to compare these two distributions directly for an evaluation of every candidate policy. Therefore, we perform this evaluation by measuring how much one dataset follows the pattern of the other by making use of the model predictions on both datasets. In detail, let us split $D_{\text{train}}$ into $D_{\mathcal{M}}$ and $D_{\mathcal{A}}$ that are used for learning the model parameter $\theta$ and exploring the augmentation policy $\mathcal{T}$, respectively. We employ the following objective to find a set of learned augmentation policies
\begin{align}
    \label{eq:objective}
    \mathcal{T}_{*} = \underset{\mathcal{T}}{\text{argmax }}\mathcal{R}(\theta^{*}|\mathcal{T}(D_{\mathcal{A}}))
\end{align}
where model parameter $\theta^{*}$ is trained on $D_{\mathcal{M}}$. It is noted that in this objective, $\mathcal{T}_{*} $ approximately minimizes the distance between density of $D_{\mathcal{M}}$ and density of $\mathcal{T}(D_{\mathcal{A}})$ from the perspective of maxmizing the performance of both model predictions with the same parameter $\theta$.

To achieve \eqref{eq:objective}, we propose an efficient strategy for augmentation policy search (see Figure \ref{fig:FAA}). First, we conduct the $K$-fold stratified shuffling \cite{shahrokh2013effect} to split the train dataset into $D_{\text{train}}^{(1)},\ldots,D_{\text{train}}^{(K)}$ where each $D_{\text{train}}^{(k)}$ consists of two datasets $D_{\mathcal{M}}^{(k)}$ and $D_{\mathcal{A}}^{(k)}$. As a matter of convenience, we omit $k$ in the notation of datasets in the remaining parts. Next, we train model parameter $\theta$ on $D_{\mathcal{M}}$ from scratch without data augmentation. Contrary to previous methods \cite{cubuk2018autoaugment, ho2019pba}, our method does not necessarily reduce the given network to child models or proxy tasks. 

After training the model parameter, for each step $1\leq t\leq T$, we explore $B$ candidate policies $\mathcal{B}=\{\mathcal{T}_{1}, \ldots, \mathcal{T}_{B}\}$ via Bayesian optimization method which repeatedly samples a sequence of sub-policies from search space $\mathcal{S}$ to construct a policy $\mathcal{T}=\{\tau_{1},\ldots,\tau_{N_{\mathcal{T}}}\}$ and tunes corresponding calling probabilities $\{p_{1},\ldots,p_{N_{\mathcal{T}}}\}$ and magnitudes $\{\lambda_{1},\ldots,\lambda_{N_{\mathcal{T}}}\}$ to minimize the expected loss $\mathcal{L}(\theta|\cdot)$ on augmented dataset $\mathcal{T}(D_{\mathcal{A}})$ (see line 6 in Algorithm \ref{alg:fast-autoaugment}). Note that, during the policy exploration-and-exploitation procedure, the proposed algorithm does not train model parameter from scratch again, hence the proposed method find augmentation policies significantly faster than AutoAugment. The concrete Bayesian optimization method is explained in Section \ref{subsubsec:optimization-density-estimation}.

As the algorithm completes the exploration step, we select top-$N$ policies over $\mathcal{B}$ and denote them $\mathcal{T}_{t}$ collectively. Finally, we merge every $\mathcal{T}_{t}$ into $\mathcal{T}_{*}$. See Algorithm \ref{alg:fast-autoaugment} for the overall procedure. At the end of the process, we augment the whole dataset $D_{\text{train}}$ with $\mathcal{T}_{*}$ and retrain the model parameter $\theta$. Through the proposed method, we can expect the performance $\mathcal{R}(\theta|\cdot)$ on augmented dataset $\mathcal{T}_{*}(D_{\mathcal{A}})$ is statistically higher than that on $D_{\mathcal{A}}$:
$$
\mathcal{R}(\theta|\mathcal{T}_{*}(D_{\mathcal{A}})) \geq \mathcal{R}(\theta|D_{\mathcal{A}})
$$
since augmentation policy $\mathcal{T}_{*}$ works as optimized inference-time augmentation \cite{simonyan2014very, szegedy2016rethinking} to make the model robustly predict correct answers. Consequently, learned augmentation policies approach \eqref{eq:objective} and improve generalization performance as we desired.

\subsubsection{Policy Exploration via Bayesian Optimization}
\label{subsubsec:optimization-density-estimation}

Policy exploration is an essential ingredient in the process of automated augmentation search. Since the evaluation of the model performance for every candidate policies is computationally expensive, we apply Bayesian optimization to the exploration of augmentation strategies. Precisely, at the line 6 in Algorithm \ref{alg:fast-autoaugment}, we employ the following Expected Improvement (EI) criterion \cite{jones2001taxonomy}
\begin{align}
    \label{eq:EI}
\text{EI}(\mathcal{T})=\mathbb{E}[\min( \mathcal{L}(\theta|\mathcal{T}(D_{\mathcal{A}}))- \mathcal{L}^{\dagger},0)]    
\end{align}
for acquisition function to explore $\mathcal{B}$ efficiently. Here, $\mathcal{L}^{\dagger}$ denotes the constant threshold determined by the quantile of observations among previously explored policies. We employ variable kernel density estimation \cite{terrell1992adaptive-kde}
on graph-structured search space $\mathcal{S}$ to approximate the criterion \eqref{eq:EI}. Practically, since the optimization method is already proposed in tree-structured Parzen estimator (TPE) algorithm \cite{bergstra2011algorithms}, we apply their \cmtt{HyperOpt} library for the parallelized implementation.




\subsection{Implementation}
\label{subsec:implementation}

\SetKwInOut{Input}{Input}
\SetKwInOut{Output}{Output}

\begin{algorithm}[t]
\DontPrintSemicolon
\Input{$\left(\theta, D_{\text{train}}, K, T, B, N \right)$}
Split $D_{\text{train}}$ into $K$-fold data $D_{\text{train}}^{(k)}=\{(D_{\mathcal{M}}^{(k)}, D_{\mathcal{A}}^{(k)})\}$ \tcp*{\cmtt{stratified shuffling}}

\For{$k \in \{1,\ldots ,K\}$}{
    $\mathcal{T}_{*}^{(k)} \leftarrow \emptyset$, $(D_{\mathcal{M}},D_{\mathcal{A}}) \leftarrow (D_{\mathcal{M}}^{(k)}, D_{\mathcal{A}}^{(k)})$ \tcp*{\cmtt{initialize}}
    Train $\theta$ on $D_{\mathcal{M}}$ \\
    \For{$t \in \{0,\ldots,T-1\}$}
    {
        $\mathcal{B} \leftarrow$ BayesOptim$(\mathcal{T},\mathcal{L}(\theta|\mathcal{T}(D_{\mathcal{A}})), B)$ \tcp*{\cmtt{explore-and-exploit}}
        $\mathcal{T}_{t}\leftarrow $ Select top-$N$ policies in $\mathcal{B}$\\
        $\mathcal{T}_{*}^{(k)} \leftarrow \mathcal{T}_{*}^{(k)} \cup \mathcal{T}_{t}$ \tcp*{\cmtt{merge augmentation policies}} 
    } 
}
\Return{$\mathcal{T}_{*} = \bigcup_{k}\mathcal{T}_{*}^{(k)}$} 
\caption{Fast AutoAugment}
\label{alg:fast-autoaugment}
\end{algorithm}

Fast AutoAugment searches desired augmentation policies applying aforementioned Bayesian optimization to distributed train splits. In other words, the overall search process consists of two steps, (1) training model parameters on $K$-fold train data with default augmentation rules and (2) exploration-and-exploitation using \cmtt{HyperOpt} to search the optimal augmentation policies. In the below, we describe the practical implementation of the overall steps in Algorithm \ref{alg:fast-autoaugment}. 
 The following procedures are mostly parallelizable, which makes the proposed method more efficient to be used in actual usage. We utilize \cmtt{Ray} \cite{ray} to implement Fast AutoAugment, which enables us to train models and search policies in a distributed manner.

\textbf{Shuffle} (Line 1): We split training sets while preserving the percentage of samples for each class (stratified shuffling) using \cmtt{StratifiedShuffleSplit} method in \cmtt{sklearn} \cite{scikit-learn}. 

\textbf{Train} (Line 4): Train models on each training split. We implement this to run parallelly across multiple machines to reduce total running time if the computational resource is enough.

\textbf{Explore-and-Exploit} (Line 6): We use \cmtt{HyperOpt} library from \cmtt{Ray} with $B$ search numbers and 20 maximum concurrent evaluations. Different from AutoAugment, we do not discretize search spaces since our search algorithm can handle continuous values. We explore one of the possible operations with probability $p$ and magnitude $\lambda$. The values of probability and magnitude are uniformly sampled from $[0,1]$ at the beginning, then \cmtt{HyperOpt} modulates the values to optimize the objective $\mathcal{L}$.

\textbf{Merge} (Line 7-9): Select the top $N$ best policies for each split and then combine the obtained policies from all splits. This set of final policies is used for re-train.

\section{Experiments and Results}
\label{sec:experiment}

In this section, we examine the performance of Fast AutoAugment on the CIFAR-10, CIFAR-100 \cite{cifar}, and ImageNet \cite{imagenet} datasets and compare the results with baseline preprocessing, Cutout \cite{devries2017cutout}, AutoAugment \cite{cubuk2018autoaugment}, and PBA \cite{ho2019pba}. For ImageNet, we only compare the baseline, AutoAugment, and Fast AutoAugment since PBA does not conduct experiments on ImageNet. We follow the experimental setting of AutoAugment for fair comparison, except that an evaluation of the proposed method on AmoebaNet-B model \cite{real2018regularized} is omitted. As in AutoAugment, each sub-policy consists of two operations ($N_{\tau} = 2$), each policy consists of five sub-policies ($N_{\mathcal{T}} = 5$), and the search space consists of the same 16 operations (\text{ShearX}, \text{ShearY}, \text{TranslateX}, \text{TranslateY}, \text{Rotate}, \text{AutoContrast}, \text{Invert}, \text{Equalize}, \text{Solarize}, \text{Posterize}, \text{Contrast}, \text{Color}, \text{Brightness}, \text{Sharpness}, \text{Cutout}, \text{Sample Pairing}). In Fast AutoAugment algorithm, we utilize 5-folds stratified shuffling ($K=5$), 2 search width ($T=2$), 200 search depth ($B=200$), and 10 selected policies ($N=10$) for policy evaluation. We increase the batch size and adapt the learning rate accordingly to boost the training \cite{you2017large}. Otherwise, we set other hyperparameters equal to AutoAugment if possible. For the unknown hyperparameters, we follow values from the original references or we tune them to match baseline performances. 

\begin{table*}[t!] \center
\begin{tabular}{c | c c c c | c }
\toprule
Model & Baseline & Cutout \cite{devries2017cutout} & AA \cite{cubuk2018autoaugment} & PBA \cite{ho2019pba} & $\underset{\text{(transfer / direct)}}{\text{Fast AA}}$ 
\tabularnewline
\midrule
Wide-ResNet-40-2 & 5.3 & 4.1 & 3.7 & $-$ & \textbf{3.6} / 3.7 \\
Wide-ResNet-28-10 & 3.9 & 3.1 & {\bf 2.6} & \textbf{2.6} & 2.7 / 2.7 \\
Shake-Shake(26 2$\times$32d) & 3.6 & 3.0 & {\bf 2.5} & \textbf{2.5} & 2.7 / \textbf{2.5} \\
Shake-Shake(26 2$\times$96d) & 2.9 & 2.6 & {\bf 2.0} & \textbf{2.0} & \textbf{2.0} / {\bf 2.0} \\
Shake-Shake(26 2$\times$112d) & 2.8 & 2.6 & {\bf 1.9} & 2.0 & 2.0 / \textbf{1.9} \\
PyramidNet+ShakeDrop & 2.7 & 2.3 & {\bf 1.5} & \textbf{1.5} & 1.8 / 1.7
\tabularnewline
\bottomrule
\end{tabular}
\caption{Test set error rate (\%) on CIFAR-10.}
\label{table1:cifar10}
\end{table*}
\begin{table*}[t!] \center
\begin{tabular}{c | c c c c | c}
\toprule
Model & Baseline & Cutout \cite{devries2017cutout} & AA \cite{cubuk2018autoaugment} & PBA \cite{ho2019pba} & $\underset{\text{(transfer / direct)}}{\text{Fast AA}}$ 
\tabularnewline
\midrule
Wide-ResNet-40-2 & 26.0 & 25.2 & 20.7 & $-$ & 20.7 / \textbf{20.6} \\
Wide-ResNet-28-10 & 18.8 & 18.4 & 17.1 & \textbf{16.7} &  17.3 / 17.3 \\
Shake-Shake(26 2$\times$96d) & 17.1 & 16.0 & {\bf 14.3} & 15.3 & 14.9 / 14.6 \\
PyramidNet+ShakeDrop & 14.0 & 12.2 & {\bf 10.7} & 10.9 & 11.9 / 11.7		
\tabularnewline
\bottomrule
\end{tabular}
\caption{Test set error rate (\%) on CIFAR-100.}
\label{table2:cifar100}
\end{table*}


\subsection{CIFAR-10 and CIFAR-100}
For both CIFAR-10 and CIFAR-100, we conduct two experiments using Fast AutoAugment: (1) direct search on the full dataset given target network (2) transfer policies found by Wide-ResNet-40-2 on the reduced CIFAR-10 which consists of 4,000 randomly chosen examples. As shown in Table 2 and 3, overall, Fast AutoAugment significantly improves the performances of the baseline and Cutout for any network while achieving comparable performances to those of AutoAugment.

\paragraph{CIFAR-10 Results}

In Table \ref{table1:cifar10}, we present the test set accuracies according to different models. We examine Wide-ResNet-40-2, Wide-ResNet-28-10 \cite{zagoruyko2016wide}, Shake-Shake \cite{gastaldi2017shake}, Shake-Drop \cite{yamada2018shakedrop} models to evaluate the test set accuracy of Fast AutoAugment. It is shown that, Fast AutoAugment achieves comparable results to AutoAugment and PBA on both experiments. We emphasize that it only takes 3.5 GPU-hours for the policy search on the reduced CIFAR-10. We also estimate the search time via full direct search. By considering the worst case, Pyramid-Net+ShakeDrop requires 780 GPU-hours which is even less than the computation time of AutoAugment (5000 GPU-hours).  

\paragraph{CIFAR-100 Results}

Results are shown in Table \ref{table2:cifar100}. Again, Fast AutoAugment achieves significantly better results than baseline and cutout. However, except Wide-ResNet-40-2, Fast AutoAugment shows slightly worse results than AutoAugment and PBA. Nevertheless, the search costs of the proposed method on CIFAR-100 are same as those on CIFAR-10. We conjecture the performance gaps between other methods and Fast AutoAugment are probably caused by the insufficient policy search in the exploration procedure or the over-training of the model parameters in the proposed algorithm.

\subsection{SVHN}

\begin{table*}[t!] \center
\begin{tabular}{c | c c c c | c}
\toprule
Model & Baseline & Cutout \cite{devries2017cutout} & AA \cite{cubuk2018autoaugment} & PBA \cite{ho2019pba} & Fast AA
\tabularnewline
\midrule
Wide-ResNet-28-10 & 1.5 & 1.3 & \textbf{1.1} & 1.2 & \textbf{1.1}
\tabularnewline
\bottomrule
\end{tabular}
\caption{Test set error rate (\%) on SVHN.}
\label{table3:svhn}
\end{table*}

We conducted an experiment with the SVHN dataset \cite{netzer2011reading} with the same settings in AutoAugment. We chose 1,000 examples randomly and applied Fast AutoAugment to find augmentation policies. The obtained policies are applied to an initial model and we obtain the comparable performance to AutoAugment. Results are shown in Table \ref{table3:svhn} and Wide-ResNet-28-10 Model with the searched policies performs better than Baseline and Cutout and it is comparable with other methods. We emphasize that we use the same settings as CIFAR while AutoAugment tuned several hyperparameters on the validation dataset. 

\subsection{ImageNet}

\begin{table*}[t!] \center
\vspace{3pt}
\begin{tabular}{c | c c | c}
\toprule
Model & Baseline & AutoAugment \cite{cubuk2018autoaugment} & Fast AutoAugment
\tabularnewline
\midrule
ResNet-50 & 23.7 / 6.9 & \textbf{22.4} / \textbf{6.2} & \textbf{22.4} / 6.3  \\
ResNet-200 & 21.5 / 5.8 & 20.00 / 5.0 & \textbf{19.4 / 4.7}
\tabularnewline
\bottomrule
\end{tabular}
\caption{Validation set Top-1 / Top-5 error rate (\%) on ImageNet.}
\label{table4:imagenet}
\end{table*}

Following the experiment setting of AutoAugment, we use a reduced subset of the ImageNet train data which is composed of 6,000 samples from randomly selected 120 classes. ResNet-50 \cite{he2016deep} on each fold were trained for 90 epochs during policy search phase, and we trained ResNet-50 \cite{he2016deep} and ResNet-200 \cite{he2016identity} with the searched augmentation policy. In Table \ref{table4:imagenet}, we compare the validation accuracies of Fast AutoAugment with those of baseline and of AutoAugment via ResNet-50 and ResNet-200. In this test, we except the AmoebaNet \cite{real2018regularized} since its exact implementation is not open to public. As one can see from the table, the proposed method outperforms benchmarks. Furthermore, our search method is 33 times faster than AutoAugment on the same experimental settings (see Table \ref{table:GPU hours}). Since extensive data augmentation protects the network from overfitting \cite{hernandez2018data}, we believe the performance will be improved by reducing the weight decay which is tuned for the model with default augmentation rules.

\section{Discussion}

 \begin{figure}
     \centering
     \includegraphics[scale=0.45]{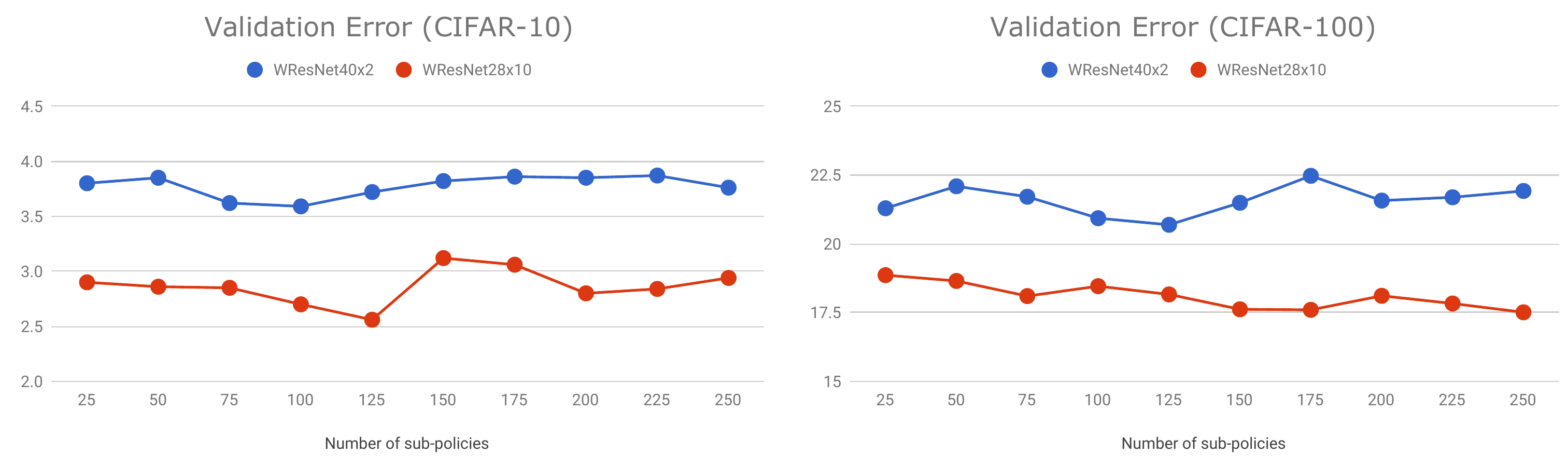}

     \caption{Validation error (\%) of Wide-ResNet-40-2 and Wide-ResNet-28-10 trained on CIFAR-10 and CIFAR-100 as number of sub-policies used in training. }
     \label{fig:number-of-policy}
 \end{figure}

Similar to AutoAugment, we hypothesize that as we increase the number of sub-policies searched by Fast AutoAugment, the given neural network should show improved generalization performance. We investigate this hypothesis by testing trained models Wide-ResNet-40-2 and Wide-ResNet-28-10 on CIFAR-10 and CIFAR-100. We select sub-policy sets from a pool of 400 searched sub-policies, and train models again with each of these sub-policy sets. Figure \ref{fig:number-of-policy} shows the relation between average validation error and the number of sub-policies used in training. This result verifies that the performance improves with more sub-policies up to 100-125 sub-policies.

As one can observe in Table \ref{table1:cifar10}-\ref{table2:cifar100}, there are small gaps between the performance of policies from direct search and the transferred policies from the reduced CIFAR-10 with Wide-ResNet-40-2. One can see that those gaps increase as the model capacities increase since the searched augmentation policies by the small model have a limitation to improve the generalization performance for the large model (e.g., Shake-Shake). Nevertheless, transferred policies are better than default augmentations; hence, one can apply those policies to different image recognition tasks.








Taking advantage of the fact that the algorithm is efficient, we experimented with searching for augmentation policies per class in CIFAR-100 with Wide-ResNet-40-2 Model. We changed search depth $B$ to 100, and kept other parameters the same. With the 70 best-performing policies per class, we obtained a slightly improved error rate of 17.2\% for  Wide-ResNet-28-10, respectively. Although it is difficult to see a definite improvement compared to AutoAugment and Fast AutoAugment, we believe that further optimization in this direction may improve performances more. Mainly, it is expected that the effect shoud be greater in the case of a dataset in which the difference between classes such as the object scale is enormous.

One can try tuning the other meta-parameters of Bayesian optimization such as search depth or kernel type in the TPE algorithm in the augmentation search phase. However, this does not significantly help to improve model performance empirically. 

\section{Conclusion}

We propose an automatic process of learning augmentation policies for a given task and a convolutional neural network. Our search method is significantly faster than AutoAugment, and its performances overwhelm the human-crafted augmentation methods. 

One can apply Fast AutoAugment to the advanced architectures such as AmoebaNet and consider various augmentation operations in the proposed search algorithm without increasing search costs. Moreover, the joint optimization of NAS and Fast AutoAugment is a a curious area in AutoML. We leave them for future works. We are also going to deal with the application of Fast AutoAugment to various computer vision tasks beyond image classification in the near future.

\newpage

\bibliographystyle{ieee}
\bibliography{bib_faa}

\end{document}